\documentclass[10pt,twocolumn,letterpaper]{article}

\usepackage{cvpr}
\usepackage{times}
\usepackage{epsfig}
\usepackage{graphicx}
\usepackage{amsmath}
\usepackage{amssymb}
\usepackage{layout}
\usepackage[nolist]{acronym} 

\usepackage{subfigure,times,multirow}
\usepackage{array}
\usepackage[dvipsnames]{xcolor}
\usepackage[ruled]{algorithm2e}
\usepackage{multirow}
\usepackage{cite}
\usepackage{kotex}
\usepackage[normalem]{ulem} 
\usepackage[keeplastbox]{flushend}
\usepackage[verbose=silent]{microtype}

\def\etal{\emph{et al}.}

\usepackage{pifont}
\newif\ifdraft\drafttrue
\ifdraft

\else

\fi

\usepackage[breaklinks=true,bookmarks=false]{hyperref}

\newcommand*\samethanks[1][\value{footnote}]{\footnotemark[#1]}

\cvprfinalcopy 

\def\cvprPaperID{3943} 

\ifcvprfinal\fi
\begin{document}
	
	\title{PPGNet: Learning Point-Pair Graph for Line Segment Detection}
	\author{
		Ziheng Zhang\thanks{Contributed equally} \quad Zhengxin Li\samethanks \quad Ning Bi \quad Jia Zheng \quad Jinlei Wang \quad Kun Huang \\ Weixin Luo \quad Yanyu Xu \quad Shenghua Gao\thanks{Corresponding author}\\
		{\small ShanghaiTech University}\\
		{\tt\scriptsize \{zhangzh, lizhx, bining, zhengjia, wangjinlei, huangkun, luowx, xuyy2, gaoshh\}@shanghaitech.edu.cn}
	}

	\maketitle
	\thispagestyle{empty}
	
	\begin{abstract}
		In this paper, we present a novel framework to detect line segments in man-made environments. Specifically, we propose to describe junctions, line segments and relationships between them with a simple graph, which is more structured and informative than end-point representation used in existing line segment detection methods. In order to extract a line segment graph from an image, we further introduce the \textit{PPGNet}, a convolutional neural network that directly infers a graph from an image. We evaluate our method on published benchmarks including \textit{York Urban} and \textit{Wireframe} datasets. The results demonstrate that our method achieves satisfactory performance and generalizes well on all the benchmarks. The source code of our work is available at \url{https://github.com/svip-lab/PPGNet}.
	\end{abstract}

	\section{Introduction}
	Retrieving 3D information from 2D images has long been a fundamental problem in computer vision. The feasibility of conventional methods based on local feature detection, matching and tracking (e.g., corners, edges, SIFT features, and patches) has been proved. However, modern applications, which involve interaction between autonomous agents and man-made physical environments, have presented more complex challenges. On the one hand, man-made environments often contain abundant homogeneous surfaces and highly repeated patterns, which introduces difficulties for feature matching and tracking. On the other hand, for some applications (e.g., visual odometry), of which the performance highly depends on the geometric primitives presenting in different views, the choice of such primitives (e.g., points, lines segments, or other structures) becomes critical: different primitives provide distinct sets of geometric information.
	
	The prior assumption about a spacial structure like \textit{Manhattan World} \cite{Delage2007Automatic,Flint2011Manhattan,Ramalingam2013Manhattan} or \textit{room topology} \cite{zou2018layoutnet,Zhao2017Physics,lee2017roomnet} could significantly benefit the 3D reconstruction, but they are often violated in real man-made environments. Instead, common junctions and line segments are capable of delivering important geometric information without dependency to any prior assumption. 
	For extensive tasks relevant to 3D vision, such as camera calibration \cite{elqursh2011line,zhang2014structure,salaun2016robust}, matching across views \cite{schmid1997automatic} and 3D reconstruction  \cite{parodi19963d,zhang2014structure,hofer2017efficient} , edges have demonstrated more robustness to lighting changes and preserve more information than points.
	Several recent works  \cite{yang2018automatic,kim2018indoor,zou2018layoutnet} show that line segments could largely facilitate 3D modeling of indoor scenes.
	
	Traditional line segment detection algorithms \cite{von2010lsd,akinlar2011edlines,lu2015cannylines,almazan2017mcmlsd} generally start from edge detection, followed by merging procedure and optionally some refinement techniques. However, such approaches are usually sensitive to changes in scale and illumination, since they merely depend on the local features.
	Additionally, some geometrically informative lines, such as intersections between two homochromatic walls, often have low local edge responses, thus tend to be ignored by such methods.
	On contrast, a human can easily recognize such visually obscure intersections through global semantic inference. 
	
	The recent success of deep learning has shown the desirable capability of image understanding, such as image classification  \cite{krizhevsky2012imagenet,szegedy2015going,he2016deep,szegedy2017inception}, object detection  \cite{girshick14CVPR,girshick2015fast,ren2015faster,He_2017_ICCV}, and semantic segmentation \cite{long2015fully,ronneberger2015u}.
	On the other hand, deep architectures are also effective in low-level tasks, such as contour detection \cite{shen2015deepcontour} and super-resolution \cite{dong2014learning}.
	\cite{huang2018learning} is a pioneer work of extracting \textit{wireframe} in man-made scenes with a deep architecture, for human-level perception of scene geometry. Their proposed network outputs pixel-wise junction confidence and directions together with a line heatmap, followed by a post-processing algorithm merging them to generate a parameterized presentation of line segments. As introduced in the literature, the conception of the \textit{wireframe} is a small subset of common line segments and junctions, which is practically defined by their dataset annotation. Considering that line segments outside the \textit{wireframe} subset also contain strong geometrical information, and line segment detection itself is still a challenging problem in computer vision, we focus on robust detection of general line segments.
	
	In this paper, we propose to describe junctions, line segments and the relationship between them with a simple graph. In our graph representation, the nodes stand for vertices and the edges stand for the connectivities between junction pairs, \ie the line segments. The graph is fully capable of describing any complex connections between junctions. Following this, we introduce the \textit{PPGNet}, a novel CNN based architecture which directly infers point-pair graph from given images. Specifically, we firstly use a backbone network for feature extraction, which is utilized to detect junctions. Then, we construct line segment candidate for each junction pair, and reuse the extracted feature to infer the connectivity of the line segment candidate. Consequently, all junctions and their connectivities are formed as a graph, which describes all line segments in the input image. It should be noted that our proposed network can predict a graph directly from a given RGB image .

	In order to train our proposed PPGNet, we need a dataset with annotated junctions as well as connectivity between every possible junction pair. However, annotations in existing datasets often ignore some overlapped line segments, thus can not be directly used to train our network. To fix this, we generate more informative graph-based annotations for existing datasets. Further, we also introduce a new large scale line segment dataset containing fully annotated indoor and outdoor samples, which fills the gap of current datasets that are either small in size for training deep architectures or lacking indoor/outdoor samples.
	
	The contribution of this work can be summarized as follow: 
	First, we introduce the new graph-based representation of line segments against commonly used endpoint representation, which is capable to describe all possible line segments in a more structured and informative way;
	second, we design a novel deep architecture that directly infers the line segment graph from the input image; third, we build a new dataset which covers both indoor and outdoor scenes with fully annotated line segments; fourth, results demonstrate that our method achieves satisfactory performance and generalizes well on multiple datasets.
	
	\section{Related Works}
	\subsection{Line segment detection}
	The mainstream pipeline of hand-crafted line segment detector generally consists of local feature extraction, pixel grouping, and optional refinement. These methods usually start from detecting pixels with high local gradient and/or edge response and then group them into line segments through iterative growing  \cite{nieto2011line}, co-linear clustering \cite{lu2015cannylines}, Hough domain accumulation \cite{matas2000robust,furukawa2003accurate,xu2015accurate} or Markov chain \cite{almazan2017mcmlsd}, \etc The line segments are optionally refined with false detection control based on the Helmholtz principle \cite{von2010lsd,akinlar2011edlines}, as well as fragment merging and endpoint relocation \cite{brown2015generalisable}. 
	
	The hand-crafted line segment detectors highly depend on carefully designated parameters. Even though some of them are parameter-free, the results still are highly sensitive to the choice of threshold. In a recent research \cite{huang2018learning}, a CNN including two branches is proposed to parse a junction map and a line heatmap from an image, which are then merged into a set of line segments. This learning-based approach outperforms the hand-crafted methods with a large margin.  Nonetheless, there is not a framework that directly outputs a parameterized presentation of line segments by far.
	
	\subsection{Junction detection}
	Although junction detection has been studied for long \cite{rosenfeld1973angle,hannah1974computer}, it remains a challenging problem. 
	A typical work is to compute local \textit{cornerness} based on so-called Harris matrix \cite{harris1988combined}, which is, however, sensitive to scale and localization.	
	Some works focus on contour curvature or continuation for detecting junctions \cite{mokhtarian1998robust,awrangjeb2008improved,cao2003good}. Other works exploit the consistency between textures and gradient-based \cite{cazorla2003two, sinzinger2008model} or pattern-based \cite{xia2014accurate, parida1998junctions} templates as an effective cue for junction detection.
	
	According to a psychophysical analysis, it is difficult to recognize junctions without context information in a large enough area, even for humans \cite{mcdermott2004psychophysics}. In this direction, \cite{maire2008using} attains robust edge and junction detection by combining local cues (e.g., lightness, color and gradient) with a \textit{global probability of boundary} (gPb) detector, which is learned from human-annotated data. Benefiting from the large receptive field of the deep neural network, \cite{huang2018learning} achieves state of the art performance on junction detection.

	\subsection{CNN-based Graph Inference}
	A convolutional neural network is capable of inferring graphs from images. In  \cite{newell2017associative}, the multi-person pose estimation problem is resolved by considering each person as a graph and grouping the body joints with associative embedding. As a more general work, a CNN is trained to detect all the objects and relationships between them in an image by means of associative embedding \cite{newell2017pixels}. The objects and relationships in a scene graph can be further refined with a gated recurrent unit (GRU) \cite{xu2017scene}.
	However, their network only outputs nodes and edges, together with embedding that associates edges to nodes, therefore extra steps are required to construct the final graph. Furthermore, their framework cannot handle arbitrary overlapped edges. In contrast, our network can infer an arbitrary simple graph parameterized by nodes and an adjacency matrix directly from the input.
	
	\section{PPGNet for Line Segment Detection}
	\subsection{Junction-line Graph Representation}
	\label{sec:problem_state}
	Here, we consider the problem of detecting line segments directly from an RGB image. We propose to use a simple graph $G^n = \{V^n, E^n\}$ to represent all line segments in a given image $X^n$ (a sample indexed by $n$ in a dataset), of which $V^n$ stands for the set of junctions and $E^n$ for the set of connectivities between any junction pairs. We now transfer the original line segment detection problem into the graph inference problem. In our implementation, $V^n$ and $E^n$ are parameterized using an ordered list of junctions $J^n$ and an adjacency matrix $A^n$, respectively. Hence, each element in $J^n$ is the coordinate of the junction and the entry $A^n_{ij}$ in the $i$-th row and $j$-th column of matrix $A^n$ equals one only if junction pair $J^n_i$ and $J^n_j$ forms a line segment.
	
\begin{figure}[ht]
		\begin{center}
			\includegraphics[width=0.47\textwidth]{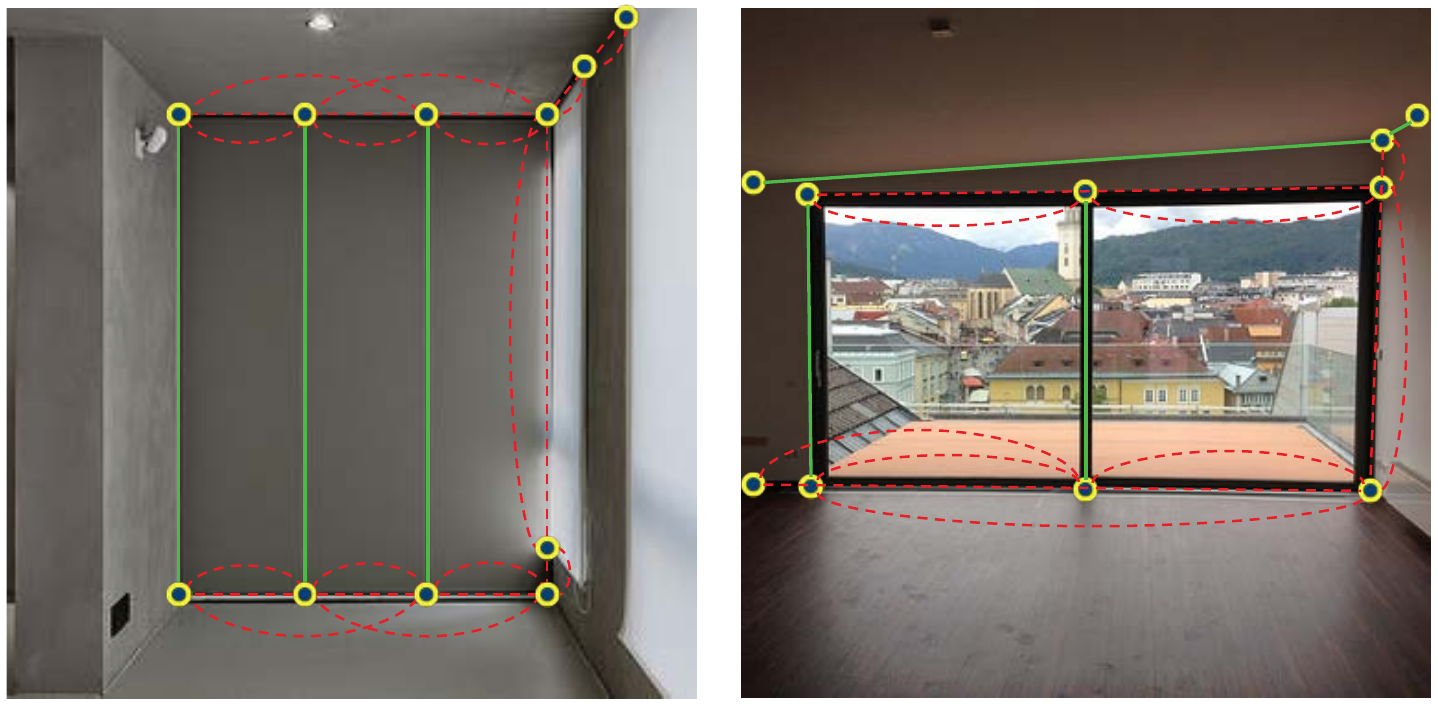}
		\end{center}
		\caption{Some cases where junctions are densely connected. The connectivities among junctions, as indicated by red dashed curves, can be more completely identified in graph representation than end-point representation. }
		\label{fig:line_compact}
	\end{figure}

	The graph representation is more structured than endpoint representation for line segments. Line segments that share the same endpoints only add more ones to the adjacency matrix without introducing extra terms. Besides, graph representation is also much more informative. Connectivities between junctions are fully described in a combinational way (Fig.\ref{fig:line_compact}) and both longer line segments and any inner shorter ones are depicted by the graph, which benefits the selection of befitting line segments from the graph in accordance with specific applications.
	
	In this work, we use a deep neural network to learn the mapping from RGB image $X$ to graph $G$. Due to the fact that $G$ fully describes all line segments in $X$ and can be transferred to endpoint representation with minor efforts, our method is a unified solution, although containing multiple stages, to settle line segment detection problem. 
	
	\begin{figure*}[ht]
		\begin{center}
			\includegraphics[width=0.85\textwidth]{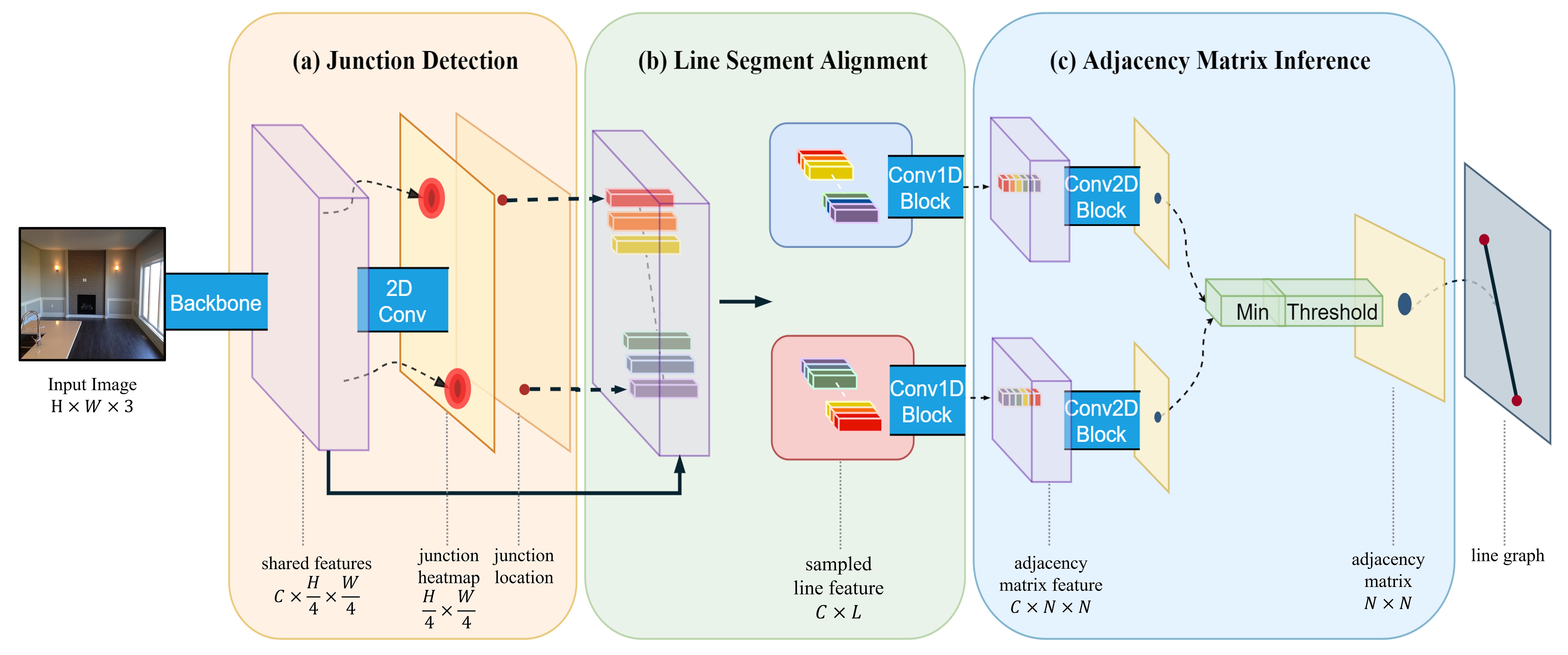}
		\end{center}
		\caption{The \textbf{PPGNet} architecture. First, the backbone computes shared features of size $C\times\frac{H}{4}\times\frac{W}{4}$ for Junction detection and adjacency matrix inference. Second, the Junction Detection Module output a list of $N$ junctions. Third, each junction pair is formed as two line segment candidates of different directions, over which features are evenly sampled into two feature matrix of size $C \times L$. After that, we apply 1D convolution over each feature matrix, which outputs a feature vector of size $C$.  Fourth, each feature vector is used by the Adjacency Matrix Inference Module to infer the connectivity of the corresponding junction pairs. }
		\label{fig:framework}
	\end{figure*}

	\subsection{PPGNet}
	Motivated by Faster R-CNN \cite{ren2015faster}, we propose a two-staged framework that detects junction points at the first stage and then identifies the connectivities between all point pairs at the second stage. The proposed PPGNet, as illustrated in Fig.\ref{fig:framework}, comprises four parts: (i) a convolutional backbone architecture for feature extraction over the entire input image, (ii) the Junction Detection Module (JDM), (iii) the Line Segment Alignment Module (LSAM) which extracts a feature tensor for the line segment candidate defined by a pair of detected junctions, and (iv) the Adjacency Matrix Inference Module (AMIM) which detects the connectivity between each junction pair. Given an image, our network predicts both junction locations and their connectivities represented by an adjacency matrix.
	
	\subsubsection{Backbone Network}
	We use the semantic segmentation network implemented by CSAIL \cite{zhou2017scene,xiao2018unified,zhou2016semantic} as our backbone network, which consists of a dilated ResNet-50 encoder and a decoder with pyramid pooling, except for the last convolution layer, of which the number of output channels $C$ is changed to be 256 instead of 1.
	\footnote{The details about the backbone can be found in the Github page\\ \url{https://github.com/CSAILVision/semantic-segmentation-pytorch}} For an input image of size $H \times W$, the backbone network extracts a 256-channel feature map of size $H/4 \times W/4$.
	
	\subsubsection{Junction Detection Module}
	The JDM extracts junctions over the input image represented by their coordinates. Unlike commonly used anchor based detection methods such as R-CNN \cite{ren2015faster}, YOLO \cite{redmon2016you} or SSD \cite{liu2016ssd}, the JDM first regresses a junction heatmap, then applies Local Maximum Filter (LMF) to get coordinates where junction response is higher than its eight neighbors. Non-maximum Suppression (NMS) is also used to avoid multiple detections of the same junctions. Unlike that in detection methods, the NMS in JDM is implemented by a hierarchical clustering using the single linkage algorithm, where the clusters are formed by the inconsistency method with a cutoff threshold (3 pixels in all our experiments).
	
	In detail, the JDM first regresses junction heatmap from the feature extracted by the backbone network through a convolutional architecture, which comprises two conv3x3-bn-relu blocks followed by a conv1x1 layer with sigmoid activation. Then it identifies all points in the heatmap where junction responses are higher than a threshold $\tau$ and are the highest among 8-neighbors. After that, the detected points are clustered into groups, within which the distance between arbitrary two points is no greater than $\epsilon$, and all the points with the highest junction responses in their groups are predicted as junction points. We use $\epsilon=3$ pixels in all our experiments.

	\subsubsection{Line Segment Alignment Module}
	Given two junctions and a feature map, the LSAM samples the feature map along the line segment candidate defined by the junction pair, and extracts a fixed-length feature vector from the feature map. LSAM works in a way similar as ROI Align Module \cite{He_2017_ICCV}, except that LSAM aligns feature vectors instead of patches.	
	
	For each junction pair and feature map of an image, the LSAM generates a feature tensor of size $C \times L$, where $C$ is the number of channels of feature map and $L$ is the spatial length of line segment feature. Specifically, LSAM first generates $L$ equidistant sampling points from the starting point to the end point of the junction pair, then uses bilinear interpolation to sample pixel value for each point on the feature map. In our main model, $L$ is set to be $64$, so that each junction pair yields a feature tensor of size $C \times 64$ for connectivity inference.
	
	\subsubsection{Adjacency Matrix Inference Module}
	The AMIM predicts the connectivity of every combination of junction pairs within an image. It takes the features for all line segment candidates provided by the LSAM, and uses a convolutional structure to predict the connection probability for each candidate.
	
	Given $K$ junctions predicted by JDM, AMIM generates a $K \times K$ adjacency matrix $A$, by which the line segment detection problem is turned into a binary classification problem of whether two junctions are connected. For every possible junction pair, two feature vectors of a line segment corresponding to different junction orders are extracted by LSAM, which are then fed into three cascaded \textit{conv2d-gn-relu} blocks, where \textit{gn} represents the Group Normalization Layer \cite{Wu_2018_ECCV}. The kernel sizes, stride size and padding size of the three convolution layers are $8, 4, 2$, respectively. After that, a single \textit{conv2d-sigmoid} block is used to get the connectivity confidence of the junction pair in different orders, of which the lowest becomes the final confidence for the junction pair. Intuitively, this processing acts as an `and' logic to ensure that a junction pair is connected regardless of the order of feature concatenation.
	
	In practice, because JDM can detect an arbitrary number of junctions, AMIM predicts one block of matrix $A$ of fixed size $64 \times 64$  at a time and runs multiple times to get the whole adjacency matrix.	Furthermore, as all possible connectivities between all junction pairs are inspected in AMIM, which causes an $O(n^2)$ complexity, it is impractical to process a too large number of junctions in AMIM. Due to an observation that JDM tends to assign a higher score to the junctions associated with more line segments, we only choose the first 512 junctions with the highest responses on the heatmap when more than 512 junctions are outputted by JDM. In our experiments, it takes about 0.9s to process an image containing 512 junctions with a Tesla P40 GPU.
	
		\begin{table*}[t]
		\caption{Dataset statistics}
		\label{tab:data_stat}
		\begin{center}
			\begin{tabular}{lccccl}
				\hline
				& \# images& resolution & \# avg. junc. & \# avg. lines & scenes$\backslash$line types \\
				\hline
				Wireframe & 5462 & $480*405(avg.)$ & 150 & 75 & indoor$\backslash$\textit{wireframe} \\
				York Urban & 102 & $640*480$ & 209 & 119 & both$\backslash$Manhattan\\
				Ours-indoor & 1378 & $900*1200$ & 67 & 41 & indoor$\backslash$general\\
				Ours-outdoor & 2534 & $2048*1080$ & 537 & 311 & outdoor$\backslash$general\\
				\hline
			\end{tabular} 
		\end{center}
	\end{table*}
	
	\subsubsection{Loss Function}
	Both junction heatmap and adjacency matrix are supervised using binary cross entropy loss, and the final loss is the weighted sum of two losses, \ie
	\begin{align*}
	&\mathcal{L} = \lambda_{junc}\mathcal{L}_{junc} + \lambda_{adj}\mathcal{L}_{adj}&\\
	&\mathcal{L}_{junc} =  -{\sum} _{i}\tilde{H}_{i}\log{H_i}+(1-\tilde{H}_{i})\log{(1-H_i)}&\\
	&\mathcal{L}_{adj} =  -{\sum} _{i}\tilde{A}_{i}\log{A_i}+(1-\tilde{A}_{i})\log{(1-A_i)}&\\
	\label{eqn:loss_function}
	\end{align*}
	where $\tilde{H}_i$ and $H_i$ are the elements of prediction and ground truth of junctions, respectively, and $\tilde{A}_i$ and $A_i$ are the elements of prediction and ground truth of the adjacency matrix, respectively. BCE stands for the cross en We set $\lambda_{junc}=\lambda_{adj}=1$ in all our experiments.
	
	\subsection{Training and Evaluating Details}
	
	All modules are jointly optimized using Stochastic Gradient Decent (SGD), with $lr=0.2$, $weight\_decay=5 \times 10^{-4}$, and $momentum=0.9$, except for all normalization layers, of which $weight\_decay$ is set to zero. The backbone network is initialized with parameters pretrained for segmentation task on the MIT ADE20K dataset, and other modules are initialized with \textit{kaiming initialization} \cite{He2015Delving}, as the common practice. During the training phase, AMIM infers adjacency matrix for ground truth junctions instead of junctions predicted by JDM because we do not have corresponding ground truth adjacency matrix for supervision. During evaluating phase, junctions and adjacency matrix are jointly estimated by our PPGNet.	

	\section{Experiments and Results}
	\begin{figure*}[t]
		\begin{center}
			\includegraphics[width=0.83\textwidth]{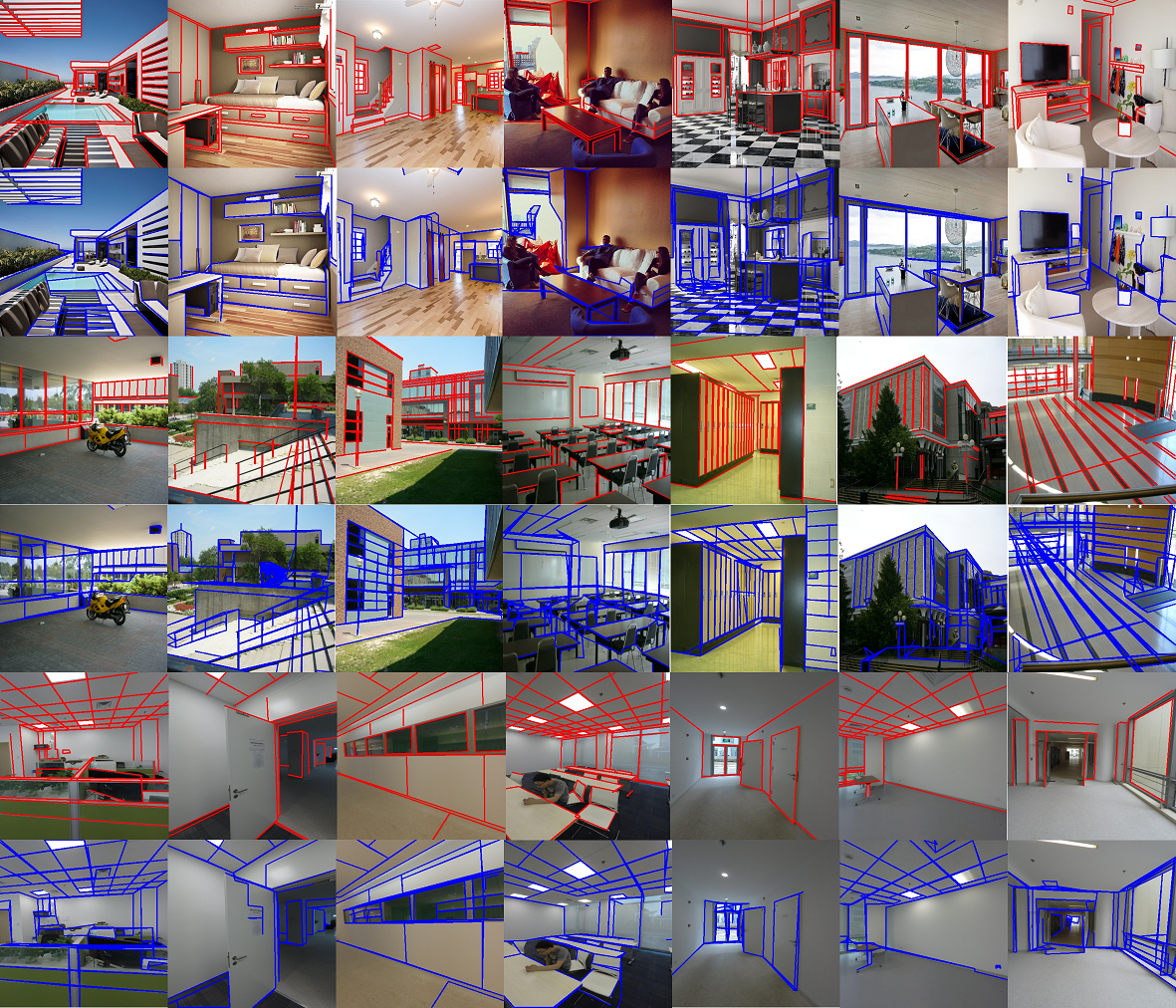}
		\end{center}
		\caption{Qualitative evaluation of our line segment detection method. 1st row: ground truth (Wireframe); 2nd row: prediction (Wireframe); 3rd row: ground truth (York Urban); 4th row: prediction (York Urban); 5th row: ground truth (Our dataset); 6th row: prediction (Our dataset)}
		\label{fig:qualitative}
	\end{figure*}
	We conduct experiments to evaluate the performance of our proposed approach and compare it with several SOTA methods. Our model is implemented with the Pytorch framework trained with four Tesla M40 GPUs. 
	\subsection{Datasets and Evaluation Metrics}
	\paragraph{Experimental Datasets}
	So far as we know, there exist two line segment datasets namely Wireframe \cite{huang2018learning} and York Urban  \cite{denis2008efficient}. However, the former only has \textit{wireframes} line segments in mostly indoor scenes annotated, while the later, though containing both indoor and outdoor scenes, is small in size (102 samples), and only has Manhattan lines labeled.
	In order to validate the capability of our framework to detect general line segments for new indoor and outdoor scenes, we build a new line segment dataset, which consists of 1,378 indoor images and 2,534 outdoor images, together with carefully labeled line segments.
	
	For the indoor part, we capture the images with resolution $900\times1200$ using a camera array that comprises seven GoPro cameras. 
	For the outdoor part, we take aerial videos of our campus with a 4K camera equipped on a DJI UAV, and extract frames with high quality at the interval of at least two seconds.
	Since the resolution of original videos is too large for labeling and training neural networks, we further crop each frame into four $2048 *1080$ images.
	
	All the line segments in both indoor and outdoor part are annotated following the protocols that any visible line segments that are longer than 10\% of the image diagonal and occluded by less than 10\% of their length is labeled. Every sample is annotated by one volunteer and then double checked by another. Different from the conception of \textit{wireframe}  \cite{huang2018learning} or Manhattan lines, the labeled line segments in our dataset only need to be visible and geometrically informative. Table \ref{tab:data_stat} summarizes the statistics of existing datasets and ours.
	
	\paragraph{Data Preprocessing}
	\label{sec:data_prep}
	In order to learn the mapping from image to line segment graph, the complete description of connectivities among all junctions is required. However, all existing datasets use endpoints to represent line segments, where both junctions and connectivities can be missing in some cases. Therefore, we introduce the data preparation scheme to convert each annotation in the original datasets into their graph representation version, which can be outlined as follows.
	
	\begin{enumerate}\setlength{\itemsep}{-\itemsep}
		\item Remove isolate junctions associated with no line segments.
		\item Find the longest line segments by searching the furthest endpoint junctions for every line segment, then mark all junctions in a longest line segment as connected. Note that the searching is not exactly along the line, but within a belt around the line, due to possible annotation error.
		\item Remove a line segment if all its inner junctions is a subset of those of another line segment. The inner junctions of a line segment are determined by their distance to the line segment.
		\item Refine each line segment by re-fitting it to its inner junctions.
		\item Refine all of the junctions that are the intersections of two or more line segments by solving the linear equations imposed by the corresponding line segments.
		\item Retrieve possibly missing junctions by finding all intersections of all line segment pairs.
		\item Construct the final line segment graph, which is parameterized by an ordered list of junctions and an adjacency matrix.
	\end{enumerate}
	
	Generally, the data preparation scheme tries to correct some bad annotations and to supplement possibly missing junctions and connectivities. We refer readers to our released code for the details of the scheme.
	
	\paragraph{Evaluation Metric}
	We quantitatively evaluate the methods using \textit{recall} and \textit{precision} as described in  \cite{maire2008using,martin2004learning,xia2014accurate,huang2018learning}. The \textit{recall} is the fraction of true line segment pixels that are detected, whereas the \textit{precision} is the fraction of detected line segment pixels that are indeed true positive. Specifically, they are calculated as follows:
	\begin{equation*}
	\mbox{Recall} \hspace{2pt} \dot{=} \hspace{2pt} |G\cap Q| / |G| 
	\mbox{, \hspace{4pt}  Precision} \hspace{2pt} \dot{=} \hspace{2pt} |G \cap Q| / |Q| \mbox{,}
	\end{equation*}
	where $G$ denotes the ground truth and $Q$ denotes the prediction. Note that, following the protocols of previous works  \cite{maire2008using,martin2004learning,xia2014accurate}, the particular measures of recall and precision allow
	for some small tolerance in the localization of line segment pixels. The tolerance in our experiments is set to be 0.01 of the image diagonal, which is the same as  \cite{huang2018learning}.
	
	\subsection{Performance evaluation}
	In order to evaluate the performance of our framework, we compared the performance of our method with the state of the art (LSD \cite{von2010lsd}, MCMLSD \cite{almazan2017mcmlsd} and \textit{wireframe} parser of Huang \etal. \cite{huang2018learning}) upon three experiment settings: (a) training and testing on Wireframe dataset using the standard splits; (b) training on Wireframe dataset and testing on the York-Urban dataset; (c) training on Wireframe dataset and testing on our dataset. Since the Wireframe and York Urban datasets are released benchmarks, we do quantitative comparison under the setting (a) and (b). Qualitative evaluation is done under the third setting to observe the generalizing capability between different data distribution.

	\paragraph{Quantitative Comparison}
	For our PPGNet, we conduct two experiments under setting (a) and (b) where (i) AMIM uses junctions predicted by JDM (marked as \textit{PPGNet}) and where (ii) AMIM uses ground truth junctions to predict line segments (marked as \textit{PPGNet*}). We include the second experiment for the reason that both the two benchmarks only have a subset of line segments annotated, \ie \textit{wireframes} and Manhattan lines, but our framework is for general line segment detection. Only with ground truth junctions can our framework understand which type of junctions should be considered.
	
	\begin{figure}[h]
		\centering
		\subfigure[]{
			\centering
			\includegraphics[width=0.22\textwidth]{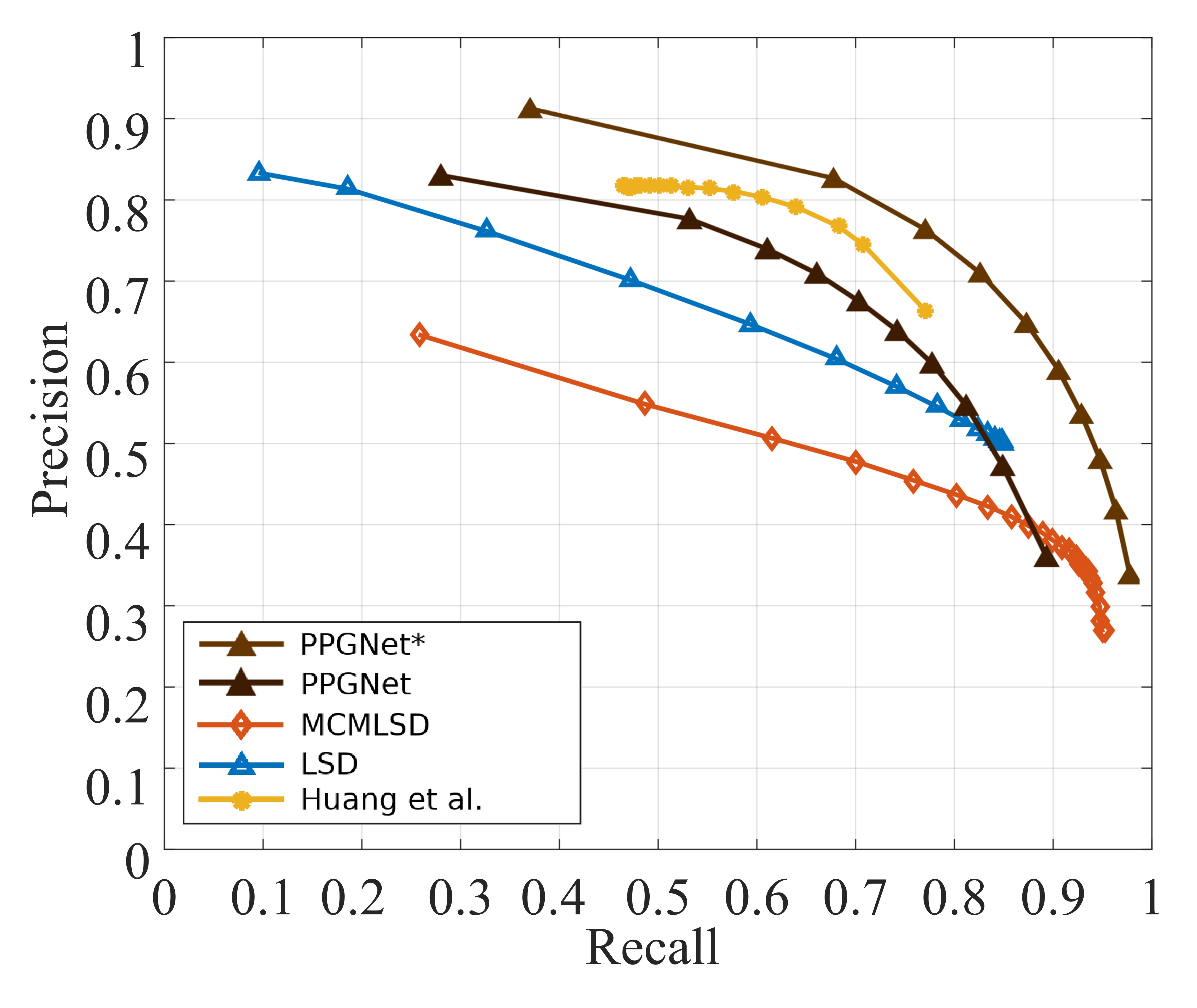}
			\label{fig:PR_exp1_sub1}
		}
		\subfigure[]{
			\centering
			\includegraphics[width=0.22\textwidth]{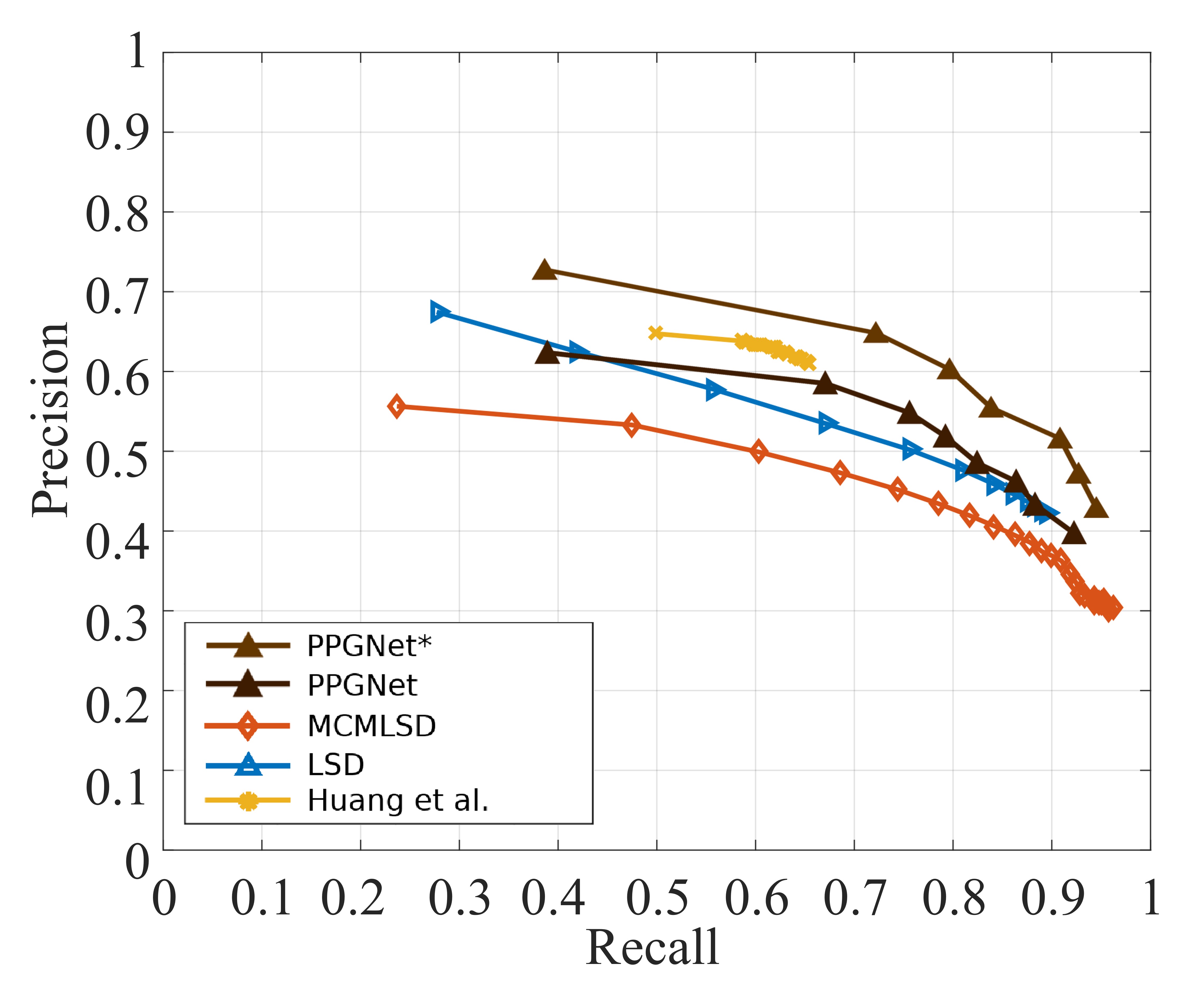}
			\label{fig:PR_exp1_sub2}
		}
		\caption{Precision-Recall curves of our PPGNet and state of the art methods evaluated on (a) Wireframe dataset and (b) York-Urban dataset. For \textit{PPGNet} and \textit{PPGNet*}, we set the threshold for JDM to $0.25$, and vary the threshold for AMIM in $[0.05, 0.1, 0.15, 0.2, 0.25, 0.3, 0.4, 0.5, 0.6, 0.7]$.}
		\label{fig:PR_curve}
	\end{figure}
	As one can see, though our PPGNet shows worse performance compared to \cite{huang2018learning} in experiment(i), it achieves superior performance in experiment(ii). In a comprehensive view, our method achieves satisfactory performance. 

	\paragraph{Qualitative Analysis}
	Fig. \ref{fig:qualitative} illustrates visualized results of line segment detection of our method on several (random) samples. It can be seen that our method is capable to robustly detect the line segments in complicated environments, and generalize well on datasets on which it has not been trained.
	
	\begin{figure*}[t]
		\begin{center}
			\includegraphics[width=0.83\textwidth]{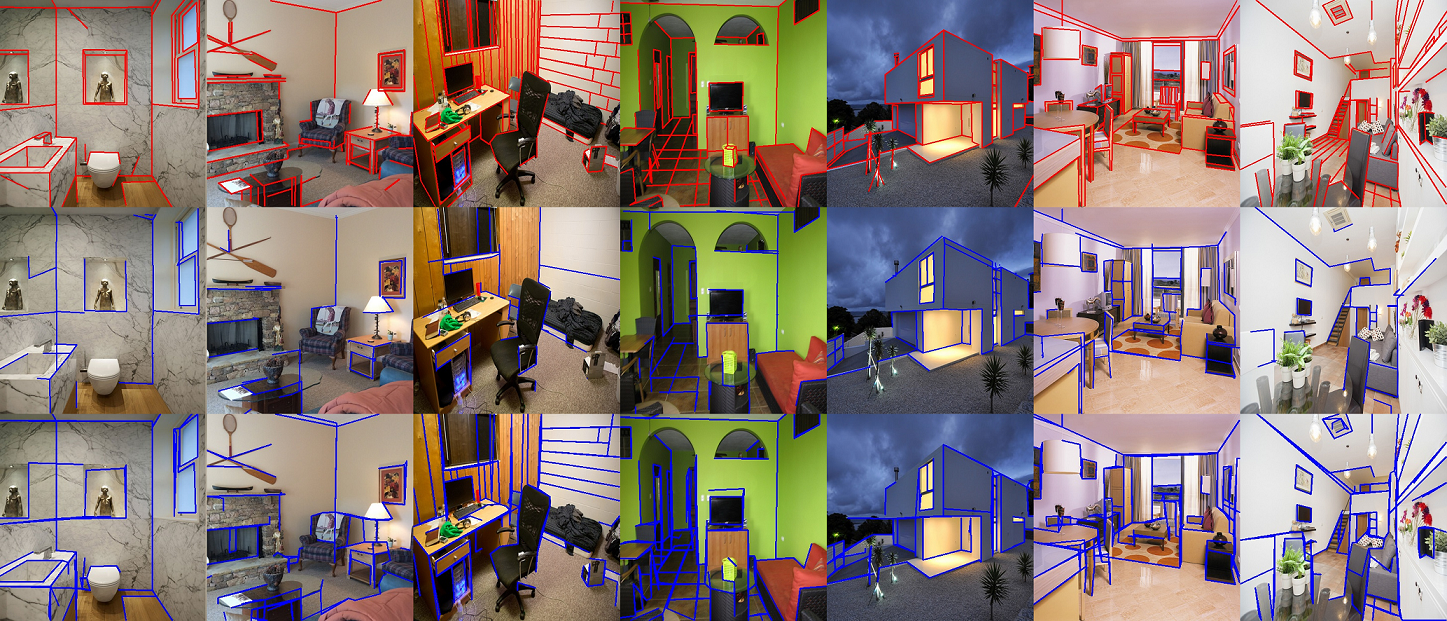}
		\end{center}
		\caption{Qualitative results on the refined \textit{Wireframe} dataset. 1st row: the ground truth; 2nd row: results of  the method proposed in \cite{huang2018learning}; 3rd row:results of \textit{PPGNet}}
		\label{fig:qual_comp}
	\end{figure*}
	
	In order to qualitatively compare the performance of our method with that of \textit{wireframe} parser \cite{huang2018learning} on general line segments, we refine the annotations of the test split in Wireframe dataset by adding the missing general line segments apart from wireframes. Fig.\ref{fig:qual_comp} shows the visualized results on some randomly picked samples for PPGNet and the method in \cite{huang2018learning}. It can be observed that our method is capable of retrieving more abundant line segment information than \cite{huang2018learning}.
	
\begin{figure}[h]
		\centering
		\subfigure[]{
			\centering
			\includegraphics[width=0.21\textwidth]{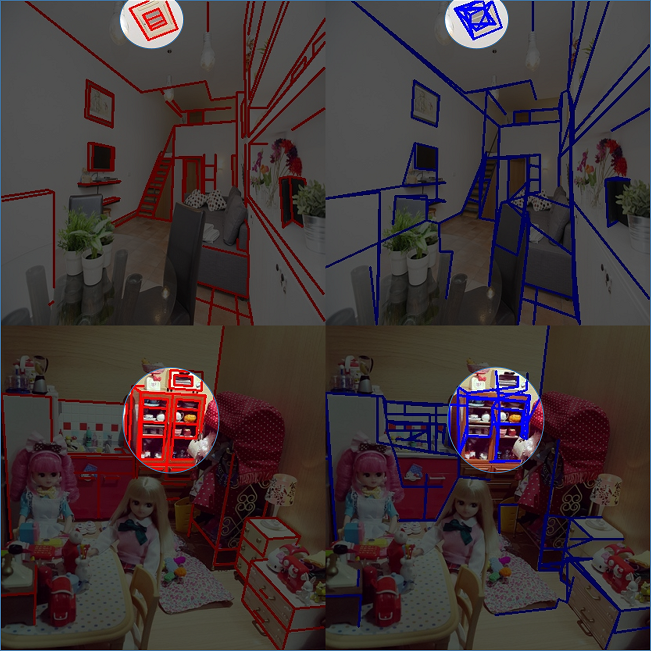}
			\label{fig:failure1}
		}\hspace{0.02\textwidth}
		\subfigure[]{
			\centering
			\includegraphics[width=0.21\textwidth]{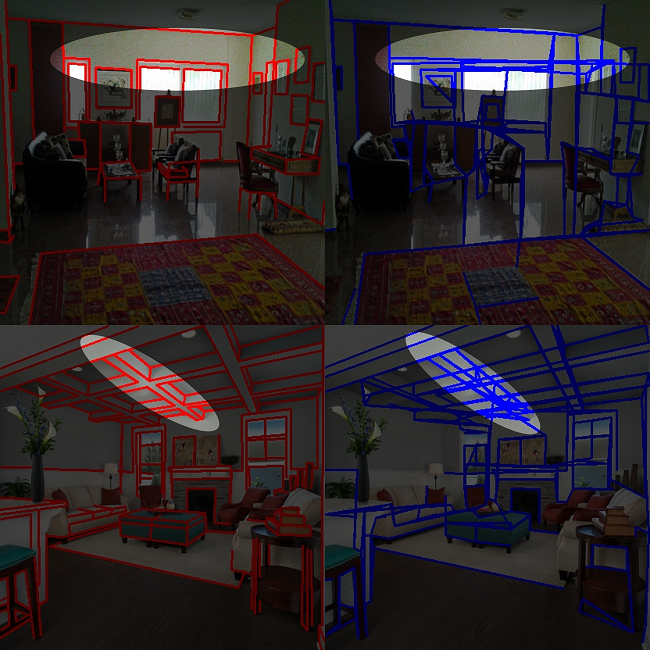}
			\label{fig:failure2}
		}
		\caption{Failure cases: (a) image samples that contain small rectangulars; (b) image samples that contain very close co-linear line segments.}
		\label{fig:failure_case}
	\end{figure}

	We also noticed that out model fails in some cases, as shown in Fig. ~\ref{fig:failure_case}. There are two typical cases: (1) for small boxes in an image, our model tends to predict the diagonal junctions as connected and (2) for close co-linear line segments, our model tends to ignore the gaps and predict all junctions as connected. These cases may be caused by the sampling procedure in AMIM. For case (1), the bilinear sampling may introduce nearby features around sampled locations. Hence the nearby junctions may interfere with connectivity prediction of current junction. On the other hand, case (2) may happen when few or no sampling points fall in the gaps between those co-linear line segments, which prevents AMIM from recognizing the discontinuity.
	
	\subsection{Junction Threshold of JDM}
\begin{figure}[h!]
		\centering
		\includegraphics[width=0.47\textwidth]{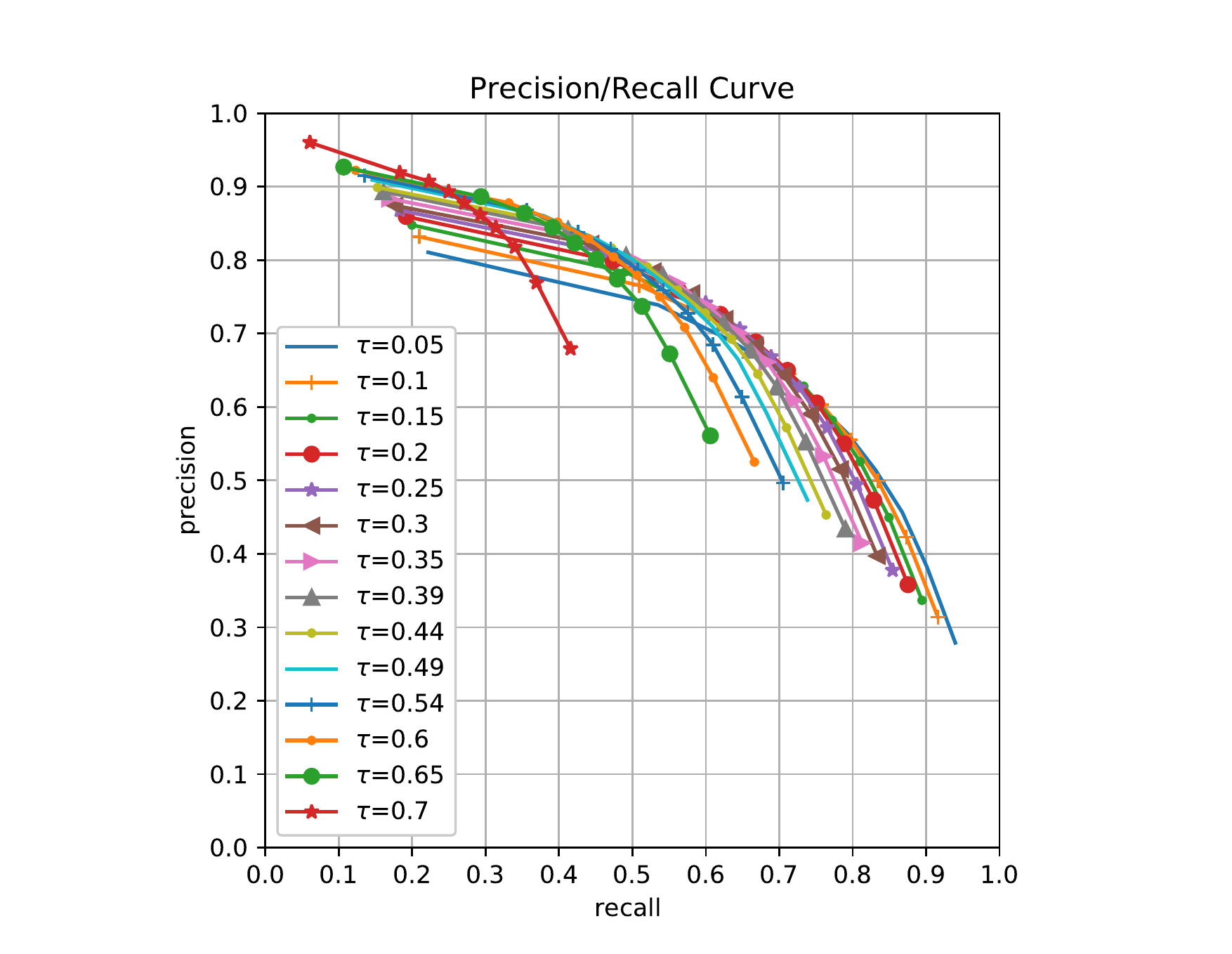}
		\caption{Precision-recall curves of different choices JDM threshold $\tau$}
		\label{fig:PR_diff_junc_thresh}
	\end{figure}

	In our framework, the AMIM predict connectivities of junctions detected by JDM, for which the threshold $\tau$ has a fundamental effect on the performance. We comprehensively compare the performance under different choices of the value for $\tau$. It can be seen in Fig.\ref{fig:PR_diff_junc_thresh} that $\tau \in [0.2, 0.3]$ lead to a better precision-recall curves. According to the quantitative evaluation in AUC, $\tau = 0.25$ is slightly better than $\tau = 0.2$ and $\tau = 0.3$.
	
	\subsection{Sampling Rate of LSAM}
\begin{figure}[h!]
		\centering
		\subfigure[]{
			\centering
			\includegraphics[width=0.28\textwidth]{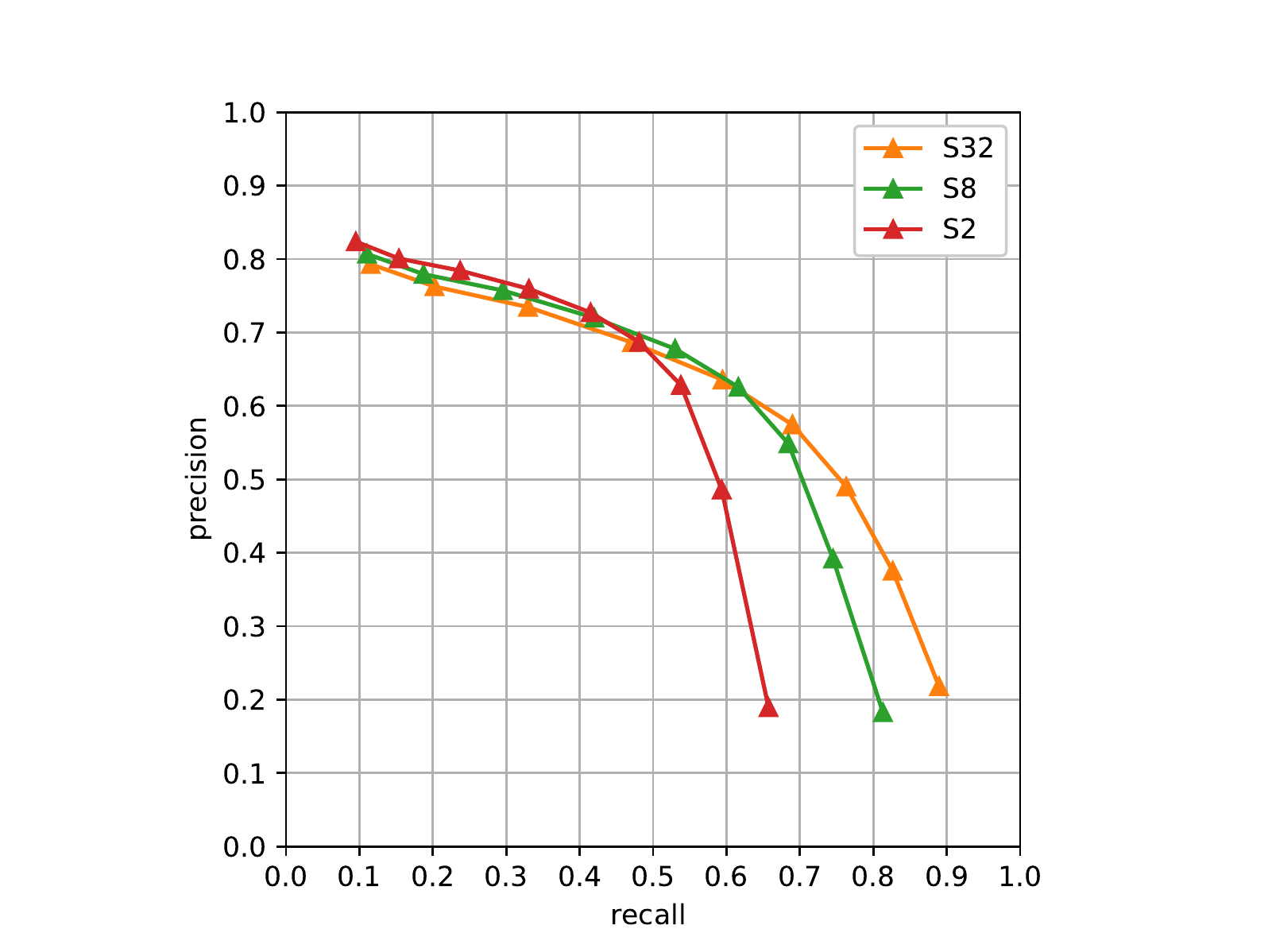}
		}
		\subfigure[]{
			\centering
			\includegraphics[width=0.17\textwidth]{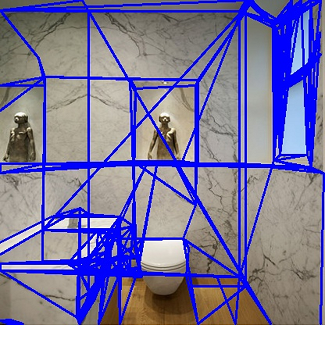}
		}
		\caption{Illustrations of (a) precision-recall curves with different sampling rate of LSAM; (b) example prediction when only junction features are pooled.}
		\label{fig:case_no_pooling}
	\end{figure}

	LSAM predicts the connectivity of junction pairs from the spatially sampled features between the two junctions. Therefore the sampling rate of LSAM has a fundamental effect on the performance of our model. In order to analyze the effect of sampling rate on performance, we conduct experiments in which different sampling rate is used in LSAM. The results are shown in Fig. \ref{fig:case_no_pooling}.(a), we can see that LSAM benefits from higher sampling rates. However, higher sampling rates also introduce extra memory usage and computational cost. Hence one should consider both performance and efficiency requirements for different applications when choosing the sampling rate.
	
	As an extreme case, the sampling rate equals 2 means that only features at the location of junction pairs are sampled. In this case, LSAM suffers from insufficient information to determine the connectivities of junction pairs. Fig.\ref{fig:case_no_pooling}.(a) shows the typical results when only two points are sampled. As one can see, LSAM fails to recognize gaps between two co-linear line segments and directions of line segments starting from the same junctions. 
	
	\section{Conclusion and Disscusion}
	In this paper, we propose to use graphs to represent all line segments in a given image and introduce the PPGNet, an multi-staged deep architecture that directly infers a graph directly from an image. Our method achieves satisfactory performance on multiple public benchmarks and shows remarkable generalization ability.
	
	There is still room for improvement in our framework. For example, currently the LSAM predicts connectivities for all possible line segments, which yields the time complexity of $O(n^2)$. Maybe one could filter some line segment candidates according to the specific applications, but there may exists a better way to further reduce the computational cost.
	
	On the other hand, PPGNet itself is a general framework to infer a graph from an image. In principle, PPGNet could also be used to solve other problems that need to detect visual parts and their spatial connections. Human pose estimation is a typical example of such problems, and we are interested in exploiting possible applications of PPGNet for such problems in future work.

	{\small
		\bibliographystyle{ieee}
		\bibliography{manuscript}
	}
	
\end{document}